# Neutrosophic information in the framework of multi-valued representation

Vasile Patrascu

Tarom Information Technology, Bucharest, Romania
e-mail: patrascu.v@gmail.com

**Abstract.** The paper presents some steps for multi-valued representation of neutrosophic information. These steps are provided in the framework of multi-valued logics using the following logical value: *true, false, neutral, unknown and saturated*. Also, this approach provides some calculus formulae for the following neutrosophic features: *truth, falsity, neutrality, ignorance, under-definedness, over-definedness, saturation and entropy*. In addition, it was defined *net truth, definedness* and *neutrosophic score*.

**Key words**: Neutrosophic information, entropy, multi-valued logic, truth, falsity, neutrality, ignorance, saturation.

## 1    Introduction

In the *fuzzy representation* of information proposed by Zadeh [23], the degree of truth is defined by $t = \mu$ while the degree of falsity is defined by $f = 1 - \mu$. There exists the partition $t + f = 1$. We extend the representation of fuzzy information to a three-valued one by defining the truth, falsity and ambiguity [17]:

$$t = \max(2\mu - 1, 0) \qquad (1.1)$$

$$f = \max(1 - 2\mu, 0) \qquad (1.2)$$

$$v = 1 - |2\mu - 1| \qquad (1.3)$$

with the partition:     $t + v + f = 1$     (1.4)

Thus we have extracted three fuzzy information features and $t \cdot f = 0$. In the *intuitionistic fuzzy representation* of information proposed by Atanassov [1], the degree of truth is defined by $t = \mu$; the degree of falsity is defined by $f = v$ while the degree of incompleteness (under-definedness) is defined by $u = 1 - \mu - v$. There exists the partition $t + u + f = 1$. We extend the representation of intuitionistic fuzzy information to a four-valued one defining the truth, falsity, ambiguity and under-definedness [17]:

$$t = \max(\mu - v, 0) \qquad (1.5)$$

$$f = \max(\nu - \mu, 0) \tag{1.6}$$

$$v = 2\min(\mu, \nu) \tag{1.7}$$

$$u = 1 - \mu - \nu \tag{1.8}$$

with the partition: $\qquad t + v + u + f = 1 \tag{1.9}$

Thus, we have extracted four fuzzy information features and $t \cdot f = 0$. The *bifuzzy information* is provided by the degree of truth $t = \mu$ and the degree of falsity $f = \nu$. Atanassov proposed a three-valued representation [2] defining the truth, falsity, and ignorance:

$$t = \mu - \frac{\min(\mu, \nu)}{2} \tag{1.10}$$

$$f = \nu - \frac{\min(\mu, \nu)}{2} \tag{1.11}$$

$$w = 1 - \max(\mu, \nu) \tag{1.12}$$

with the partition: $\qquad t + w + f = 1 \tag{1.13}$

There exist two variants for tetra-valued representation of bifuzzy information. The first was obtained, defining the truth, falsity, under-definedness and over-definedness [12], [13]:

$$t = \min(\mu, 1 - \nu) \tag{1.14}$$

$$f = \min(1 - \mu, \nu) \tag{1.15}$$

$$u = \max(1 - \mu - \nu, 0) \tag{1.16}$$

$$o = \max(\mu + \nu - 1, 0) \tag{1.17}$$

with the partition:

$$t + o + u + f = 1 \tag{1.18}$$

and having the property: $u \cdot o = 0$. The second was obtained, defining the truth, falsity, ignorance and contradiction [5], [6]:

$$t = \max(\mu - \nu, 0) \tag{1.19}$$

$$f = \max(\nu - \mu, 0) \tag{1.20}$$

$$w = 1 - \max(\mu, \nu) \tag{1.21}$$

$$c = \min(\mu, \nu) \tag{1.22}$$

with the fuzzy partition:

$$t + c + w + f = 1 \tag{1.23}$$

and having the property: $t \cdot f = 0$. In [15], [17] and [18], it is proposed a penta-valued representation, defining the truth, falsity, under-definedness, over-definedness and ambiguity (vagueness):

$$t = \max(\mu - \nu, 0) \tag{1.24}$$

$$f = \max(\nu - \mu, 0) \tag{1.25}$$

$$u = \max(1 - \mu - \nu, 0) \tag{1.26}$$

$$o = \max(\mu + \nu - 1, 0) \tag{1.27}$$

$$v = 1 - |\mu - \nu| - |\mu + \nu - 1| \tag{1.28}$$

with the fuzzy partition:

$$t + o + v + u + f = 1 \tag{1.29}$$

and having the property: $u \cdot o = 0$, $t \cdot f = 0$

The *neutrosophic representation* of information was proposed by Smarandache [3], [20], and [21]. This representation is defined by degree of truth T, degree of falsity F and degree of neutrality I. This paper will redefine for the neutrosophic information the features presented above like truth, falsity, ignorance, over-definedness, under-definedness, net truth, definedness, entropy and it introduces new features like neutrality, saturation and neutrosophic index. In the following, the paper has the structure: Section 2 presents the construction of net truth, definedness, neutrosophic score; Section 3 presents two calculus formulae for neutrosophic entropy; Section 4 presents a tetra-valued representation based on truth, falsity, neutrality and ignorance; Section 5 presents a penta-valued representation of neutrosophic information based on truth, falsity, neutrality, saturation and ignorance; Section 6 presents a penta-valued one based on truth, falsity, neutrality, over-definedness and under-definedness. Section 7 outlines some conclusions.

## 2    The Net Truth, Definedness and Neutrosophic Score

We will extend the notion of net truth and definedness from bifuzzy information to neutrosophic one. For bifuzzy information [18] it was defined the net truth and definedness. We will extend using the third component neutrality $I$ in the following way:

$$\tau = \frac{T - F}{1 + I} \tag{2.1}$$

In the next we will construct a similar function for neutrosophic definedness. We will denote the mean of neutrosophic component:

$$\lambda = \frac{T + I + F}{3} \qquad (2.2)$$

The neutrosophic definedness is described by a function: $\omega : [0,1] \to [-1,1]$ having the following properties:

$$\omega(0) = -1 , \ \omega\left(\frac{1}{3}\right) = 0 , \ \omega(1) = 1 \text{ and } \omega \text{ increases with its argument.}$$

Here are some examples:

$$\omega(\lambda) = \frac{3\lambda - 1}{1 + \lambda} \qquad (2.3)$$

$$\omega(\lambda) = 2\sin\left(\lambda \frac{\pi}{2}\right) - 1 \qquad (2.4)$$

$$\omega(\lambda) = \frac{7\lambda - 3\lambda^2}{2} - 1 \qquad (2.5)$$

$$\omega(\lambda) = \frac{9\lambda - 3 - |3\lambda - 1|}{4} \qquad (2.6)$$

$$\omega(\lambda) = \frac{\sqrt{2\lambda} - \sqrt{1 - \lambda}}{\sqrt{2\lambda} + \sqrt{1 - \lambda}} \qquad (2.7)$$

If the neutrosophic definedness is positive, the information is inconsistent or over-defined, if it is zero, the neutrosophic information is consistent or complete and if it is negative, the information is incomplete or under-defined. The pair $(\tau, \omega)$ provides a bi-valued representation of neutrosophic information. Combining these two components into a scalar, one obtains the neutrosophic score or neutrosophic index defined by:

$$\eta = \frac{\tau}{1 + |\omega|} \qquad (2.8)$$

The neutrosophic index defines the following order relation:

$$(T_1, I_1, F_1) \geq (T_2, I_2, F_2) \quad \Leftrightarrow \quad \eta_1 \geq \eta_2$$

## 3    The Neutrosophic Entropy

For neutrosophic entropy, we will trace the Kosko idea for fuzzy entropy calculus [8]. Kosko proposed to measure the entropy by a similarity function between the distance to the nearest crisp element and the distance to the farthest crisp element. For neutrosophic information the two crisp elements are $(1,0,0)$ and $(0,0,1)$. We consider the following vector: $V = (T - F, T + I + F - 1, I)$. For $(1,0,0)$ and $(0,0,1)$ it results: $V_T = (1,0,0)$ and $V_F = (-1,0,0)$. We will compute the distances:

$$D(V, V_T) = |T - F - 1| + |T + I + F - 1| + I \qquad (3.1)$$

$$D(V, V_F) = |T - F + 1| + |T + I + F - 1| + I \qquad (3.2)$$

The entropy will be defined by the similarity between these two distances. Using the Czekanowskyi formula [4], [10] it results the similarity and the entropy:

$$S_C = 1 - \frac{|D(V,V_T) - D(V,V_F)|}{D(V,V_T) + D(V,V_F)} \qquad (3.3)$$

$$E_C = 1 - \frac{|T-F|}{1 + I + |T+I+F-1|} \qquad (3.4)$$

For $T + I + F = 1$, it results the intuitionistic fuzzy entropy proposed by Patrascu [19]. For $T + F = 1$ and $I = 0$, it results the fuzzy entropy proposed by Kaufman [7].

Using the Ruzicka formula [4], [11] it results the similarity and entropy formulae:

$$S_R = 1 - \frac{|D(V,V_T) - D(V,V_F)|}{\max(D(V,V_T), D(V,V_F))} \qquad (3.5)$$

$$E_R = \frac{1 - |T-F| + I + |T+I+F-1|}{1 + |T-F| + I + |T+I+F-1|} \qquad (3.6)$$

For $I = 0$, it result the bifuzzy entropy formula [18]. If $T + I + F = 1$, it results the intuitionistic fuzzy entropy proposed by Szmidt and Kacprzyk [22]. For $T + F = 1$ and $I = 0$, it results the fuzzy entropy proposed by Kosko [8].

## 4    Tetra-valued representation of neutrosophic information

Formulae (1.10-1.12) compress the unit square to a triangle with the vertices (1,0), (0,1) and (0,0). Here we will extend this compression from 2-dimensional space to 3-dimensional one. We compress the unit cube to a tetrahedron with the vertices (1,0,0), (0,1,0), (0,0,1) and (0,0,0). We define the truth, falsity, neutrality and ignorance:

$$t = T - \frac{\min(T,I) + \min(T,F)}{2} + \frac{\min(T,I,F)}{3} \qquad (4.1)$$

$$f = F - \frac{\min(F,I) + \min(T,F)}{2} + \frac{\min(T,I,F)}{3} \qquad (4.2)$$

$$n = I - \frac{\min(T,I) + \min(I,F)}{2} + \frac{\min(T,I,F)}{3} \qquad (4.3)$$

$$w = 1 - \max(T,I,F) \qquad (4.4)$$

These four parameters define a partition of unity.

$$t + f + n + w = 1 \qquad (4.5)$$

Having this representation, the neutrosophic information could be *true, false, neutral or unknown*. These four information features have the following prototype:

T=(1,0,0); F=(0,0,1); N=(0,1,0); W=(0,0,0). For this tetra-valued representation the *indeterminacy* has two components: *neutrality* and *ignorance*, namely:

$$i = n + w \qquad (4.6)$$

The method presented in this section provides a way to transform any neutrosophic information into an incomplete one, because from (4.5) it results:

$$t + f + n \leq 1 \qquad (4.7)$$

We must mention that, for $I = 0$, one obtains (1.10), (1.11) and (1.12) proposed by Atanassov for transforming a bifuzzy set into an intuitionistic one [2]. We define the union, intersection and the negation.

*The Negation*:

For any $q = (t, n, w, f)$ the negation is defined by:

$$\bar{q} = (f, n, w, t) \qquad (4.8)$$

For the union and intersection we will use the formulae proposed in [16]. The formulae proposed here are different from those proposed by Ashbacher [3].

*The Union*

For $q_1 = (t_1, n_1, w_1, f_1)$ and $q_2 = (t_2, n_2, w_2, f_2)$

$$t_{q_1 \cup q_2} = t_1 \vee t_2 \qquad (4.9)$$

$$n_{q_1 \cup q_2} = (n_1 + t_1) \vee (n_2 + t_2) - t_1 \vee t_2 \qquad (4.10)$$

$$w_{q_1 \cup q_2} = (w_1 + f_1) \wedge (w_2 + f_2) - f_1 \wedge f_2 \qquad (4.11)$$

$$f_{q_1 \cup q_2} = f_1 \wedge f_2 \qquad (4.12)$$

*The Intersection*:

$$t_{q_1 \cap q_2} = t_1 \wedge t_2 \qquad (4.13)$$

$$w_{q_1 \cap q_2} = (w_1 + t_1) \wedge (w_2 + t_2) - t_1 \wedge t_2 \qquad (4.14)$$

$$n_{q_1 \cap q_2} = (n_1 + f_1) \vee (n_2 + f_2) - f_1 \vee f_2 \qquad (4.15)$$

$$f_{q_1 \cap q_2} = f_1 \vee f_2 \qquad (4.16)$$

where $\wedge$ represents any Frank t-norm [9] and $\vee$ represents its t-conorm. The above remark is valid for all the next section of this paper where the symbols $\wedge, \vee$ are used.

# 5 Penta-valued Representation of Neutrosophic Information Based on Subtracting of Saturation

We will extend the tetra-valued representation presented in the section 3, adding the index of saturation. In order to not make confusion, in this section we will rename the parameter defined by formulae (4.1), (4.2) and (4.3) by:

$$\beta_t = T - \frac{\min(T, I) + \min(T, F)}{2} + \frac{\min(T, I, F)}{3} \qquad (5.1)$$

$$\beta_f = F - \frac{\min(F, I) + \min(T, F)}{2} + \frac{\min(T, I, F)}{3} \qquad (5.2)$$

$$\beta_n = T - \frac{\min(T,I) + \min(I,F)}{2} + \frac{\min(T,I,F)}{3} \qquad (5.3)$$

Now, we will define the index of saturation:

$$s = 3\min(\beta_t, \beta_f) \qquad (5.4)$$

We will redefine the truth, falsity, neutrality by:

$$t = \beta_t - \min(\beta_t, \beta_f) \qquad (5.5)$$

$$f = \beta_f - \min(\beta_t, \beta_f) \qquad (5.6)$$

$$n = \beta_f - \min(\beta_t, \beta_f) \qquad (5.7)$$

with the following equivalent form:

$$s = 1 - \min(T, F, I) \qquad (5.8)$$

$$t = T - \frac{\min(T,I) + \min(T,F)}{2} \qquad (5.9)$$

$$f = F - \frac{\min(F,I) + \min(T,F)}{2} \qquad (5.10)$$

$$n = T - \frac{\min(T,I) + \min(I,F)}{2} \qquad (5.11)$$

These five parameters verify the following condition:

$$t + f + n + s + w = 1 \qquad (5.14)$$

Having this representation, the neutrosophic information could be *true, false, neutral, saturated* and *unknown*. These five information features have the following prototype: T=(1,0,0); F=(0,0,1); N=(0,1,0); S=(1,1,1); W=(0,0,0). For this penta-valued representation the *indeterminacy* has three components: *neutrality, saturation* and *ignorance*, namely:

$$i = n + s + w \qquad (5.15)$$

We define the union, intersection and the negation.

*The Negation*:

For any $q = (t, n, s, w, f)$ the negation is defined by:

$$\overline{q} = (f, n, s, w, t) \qquad (5.16)$$

For the union and intersection we will use the formulae proposed in [14].

*The Union*

For $q_1 = (t_1, n_1, s_1, w_1, f_1)$ and $q_2 = (t_2, n_2, s_2, w_2, f_2)$

$$t_{q_1 \cup q_2} = t_1 \vee t_2 \qquad (5.17)$$

$$s_{q_1 \cup q_2} = (s_1 + t_1) \vee (s_2 + t_2) - t_1 \vee t_2 \qquad (5.18)$$

$$n_{q_1 \cup q_2} = 1 - t_{q_1 \cup q_2} - f_{q_1 \cup q_2} - s_{q_1 \cup q_2} - w_{q_1 \cup q_2} \qquad (5.19)$$

$$w_{q_1 \cup q_2} = (w_1 + f_1) \wedge (w_2 + f_2) - f_1 \wedge f_2 \qquad (5.20)$$

$$f_{q_1 \cup q_2} = f_1 \wedge f_2 \qquad (5.21)$$

*The Intersection*:

$$t_{q_1 \cap q_2} = t_1 \wedge t_2 \qquad (5.22)$$

$$w_{q_1 \cap q_2} = (w_1 + t_1) \wedge (w_2 + t_2) - t_1 \wedge t_2 \qquad (5.23)$$

$$n_{q_1 \cap q_2} = 1 - t_{q_1 \cap q_2} - f_{q_1 \cap q_2} - s_{q_1 \cap q_2} - w_{q_1 \cap q_2} \qquad (5.24)$$

$$s_{q_1 \cap q_2} = (s_1 + f_1) \vee (s_2 + f_2) - f_1 \vee f_2 \qquad (5.25)$$

$$f_{q_1 \cap q_2} = f_1 \vee f_2 \qquad (5.26)$$

# 6    Penta-valued Representation Based on Neutrosophic Definedness

We denote by $\omega^+ = \max(\omega, 0)$ and $\omega^- = \max(-\omega, 0)$. Using the neutrosophic definedness we define the truth, falsity, neutrality, over-definedness and under-definedness by:

$$t = \frac{(1 - \omega^+)}{3\lambda + \omega^-} \cdot T \qquad (6.1)$$

$$n = \frac{(1 - \omega^+)}{3\lambda + \omega^-} \cdot I \qquad (6.2)$$

$$f = \frac{(1 - \omega^+)}{3\lambda + \omega^-} \cdot F \qquad (6.3)$$

$$o = \omega^+ \qquad (6.4)$$

$$u = \frac{\omega^-}{3\lambda + \omega^-} \qquad (6.5)$$

These five parameters verify the condition of fuzzy partition, namely:

$$t + n + f + o + u = 1 \qquad (6.6)$$

Having this representation, the neutrosophic information could be *true, false, neutral, over-defined* and *under-defined*. For this penta-valued representation the *indeterminacy* has three components: *neutrality, over-definedness* and *under-definedness*, namely:

$$i = n + o + u \qquad (6.7)$$

We must draw attention to the difference between *saturation* that represents the similarity to the vector $(1,1,1)$ and the *over-definedness* that is related to the inequality $T + I + F > 1$. In the same time, for both parameters, the maximum is obtained for $T = F = I = 1$. Also, the *ignorance* supplies a similarity to the vector $(0,0,0)$ while the *under-definedness* represents a measure of the inequality $1 > T + I + F$.

We define the union, intersection and the negation.

*The Negation*:

For any $q = (t, n, o, u, f)$ the negation is defined by:

$$\overline{q} = (f, n, o, u, t) \qquad (6.8)$$

For the union and intersection we will use the formulae proposed in [15].

*The Union*

For $q_1 = (t_1, n_1, o_1, u_1, f_1)$ and $q_2 = (t_2, n_2, o_2, u_2, f_2)$

$$t_{q_1 \cup q_2} = t_1 \vee t_2 \tag{6.9}$$

$$n_{q_1 \cup q_2} = 1 - t_{q_1 \cup q_2} - f_{q_1 \cup q_2} - o_{q_1 \cup q_2} - u_{q_1 \cup q_2} \tag{6.10}$$

$$o_{q_1 \cup q_2} = (o_1 + f_1) \wedge (o_2 + f_2) - f_1 \wedge f_2 \tag{6.11}$$

$$u_{q_1 \cup q_2} = (u_1 + f_1) \wedge (u_2 + f_2) - f_1 \wedge f_2 \tag{6.12}$$

$$f_{q_1 \cup q_2} = f_1 \wedge f_2 \tag{6.13}$$

*The Intersection*:

$$t_{q_1 \cap q_2} = t_1 \wedge t_2 \tag{6.14}$$

$$n_{q_1 \cap q_2} = 1 - t_{q_1 \cap q_2} - f_{q_1 \cap q_2} - o_{q_1 \cap q_2} - u_{q_1 \cap q_2} \tag{6.15}$$

$$o_{q_1 \cap q_2} = (o_1 + t_1) \wedge (o_2 + t_2) - t_1 \wedge t_2 \tag{6.16}$$

$$u_{q_1 \cap q_2} = (u_1 + t_1) \wedge (u_2 + t_2) - t_1 \wedge t_2 \tag{6.17}$$

$$f_{q_1 \cap q_2} = f_1 \vee f_2 \tag{6.18}$$

## 7 Conclusion

Multi-valued representation of neutrosophic information is presented in the paper, mainly in order to model features of its certainty and uncertainty. The proposed representations verify the condition of fuzzy partition and are accompanied by operators like negation, union and intersection. It was extended the concepts of certainty like truth and falsity, the concepts of uncertainty like ignorance, ambiguity, over-definedness, under-definedness, entropy. In addition, it was defined new concepts related to the particularity of neutrosophy like saturation, neutrality and neutrosophic score. The particularization of the obtained formulae for neutrosophic information leads to some formulae that already exist in the specialty literature and are related to intuitionistic fuzzy information and fuzzy one. This fact proves the effectiveness of our approach.